\pdfoutput=1

\documentclass[11pt]{article}

\usepackage{ACL2023}

\usepackage{times}
\usepackage{latexsym}

\usepackage[T1]{fontenc}

\usepackage[utf8]{inputenc}

\usepackage{microtype}

\usepackage{inconsolata}
\usepackage{graphicx}%
\usepackage{tabularray}
\usepackage{multirow}
\usepackage{makecell}
\usepackage{amsmath}
\usepackage{enumitem}
\usepackage{booktabs}
\usepackage{xcolor}
\definecolor{mint}{RGB}{50,255,255}
\usepackage{float}
\usepackage{csquotes}
\usepackage{bm}

\makeatletter
\def\hlinewd#1{%
\noalign{\ifnum0=`}\fi\hrule \@height #1 %
\futurelet\reserved@a\@xhline}
\makeatother

\title{Unlocking Structure Measuring: Introducing PDD, \\an Automatic Metric for Positional Discourse Coherence}

\author{
 \textbf{Yinhong Liu}$^\spadesuit$  \quad \quad
\textbf{Yixuan Su}$^\spadesuit$  \quad \quad 
 \textbf{Ehsan Shareghi}$^{\heartsuit \spadesuit}$ \quad \quad 
 \textbf{Nigel Collier}$^\spadesuit$  \quad
 \\
 $^\spadesuit$Language Technology Lab, University of Cambridge\\
 $^\heartsuit$Department of Data Science and AI, Monash University\\
 {\tt \{yl535,ys484,nhc30\}@cam.ac.uk}\\
 {\tt ehsan.shareghi@monash.edu}
}

\begin{document}
\maketitle
\begin{abstract}
Recent large language models~(LLMs) have shown remarkable performance in aligning generated text with user intentions across various tasks. 
When it comes to long-form text generation, there has been a growing interest in generation from a discourse coherence perspective.
However, existing lexical or semantic metrics such as BLEU, ROUGE, BertScore cannot effectively capture the discourse coherence.
The development of discourse-specific automatic evaluation methods for assessing the output of LLMs warrants greater focus and exploration. 
In this paper, we present a novel automatic metric designed to quantify the discourse divergence between two long-form articles.
Extensive experiments on three datasets from representative domains demonstrate that our metric aligns more closely with human preferences and GPT-4 coherence evaluation, outperforming existing evaluation methods.
\footnote{Our code is available at \url{https://github.com/williamLyh/pos_div_metric}}

\end{abstract}

\section{Introduction}

\begin{figure}[t]
    \centering
    \includegraphics[scale=0.85]{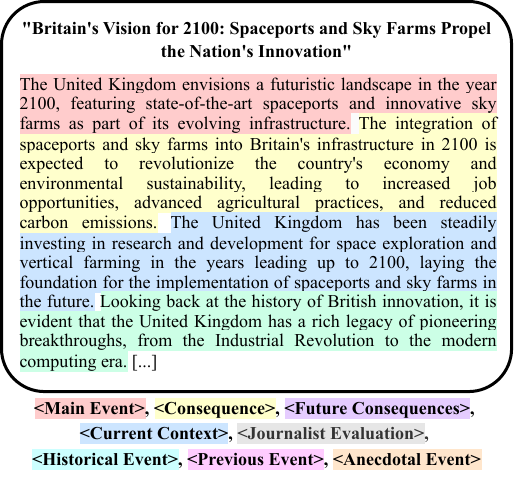}  
    \caption{A news article example with discourse role annotations. The discourse schema follows the News discourse theory by \citet{van2013news}.}
    \label{fig:news_example}
    \vspace{-5mm}
\end{figure}

Real-life texts often exhibit underlying structures.
News articles, for instance, adhere to a specific narrative order, as illustrated in Fig.~\ref{fig:news_example}, employed by journalists to efficiently convey messages and improve reader experience. 
Despite recent advances in generation of fluent text, 
\citet{deng-etal-2022-model} demonstrate that transformer-based models struggle to effectively capture and learn the underlying latent transition structure of coherent text.
Consequently, generating structurally coherent text remains an under-explored area of research.
Follow the theory of functional discourse structure, elaborated in Appendix~\ref{appen:discourse structure}, we leverage the discourse structure to model the coherence of long-form texts. 
Several recent works~\citep{spangher-etal-2022-sequentially, liu-etal-2022-plug} have addressed the problems of generating long-form text while following specific in-domain discourse schema.

While established automatic metrics such as BLEU~\citep{papineni2002bleu}, ROUGE-L~\citep{lin2002manual} and BertScore~\citep{zhang2019bertscore} exist for Natural Language Generation evaluation, they predominantly measure lexical n-gram overlaps or semantic similarities.
The evaluation of structural coherence has been a long-existing challenge \citep{guan-etal-2021-long, cho-etal-2019-towards, zhu-bhat-2020-gruen,deng-etal-2022-model}. 
A common baseline metric for measuring functional discourse structure is the exact match, which compares structure elements one-to-one at each exact position.
However, this metric is notably sensitive to local variations and differences in the lengths of articles.

To address this gap, we propose a novel automatic model-free metric, Positional Discourse Divergence (PDD), specifically designed to evaluate the underlying discourse structure of articles in comparison to references.
PDD partitions the sentences of an article into multiple position bins and calculates the divergence in discourse structures within each bin.
This approach renders PDD resilient to various challenges encountered in long-form text generation, such as accommodating local variations and handling misaligned numbers of sentences.

To validate the effectiveness and generalizability of the PDD, we evaluate the inter-agreement with human evaluations and GPT-4 coherence evaluations on three representative datasets with different discourse schema: News Discourse~\citep{choubey-etal-2020-discourse}, Long-Form Question Answering~\citep{xu-etal-2022-answer} and Recipe1M+~\citep{liu-etal-2022-plug}.
Across all three domains, PDD demonstrates the highest agreement with human judgements on coherence.

\section{Positional Discourse Divergence}

Texts within a specific genre often exhibit similar patterns in their discourse sequences, albeit with some variations at a local level.
In other words, the distribution of discourse roles is inherently tied to their approximate positions within the articles.
For instance, News reports commonly present main events and their consequences at the beginning to capture the reader's interest, even though the precise order can differ. Likewise, recipes tend to follow a predictable structure, where the preparation of ingredients is generally mentioned first, followed by cooking actions towards the middle or end of the text.

\begin{figure}
    \centering
    \includegraphics[width=\linewidth]{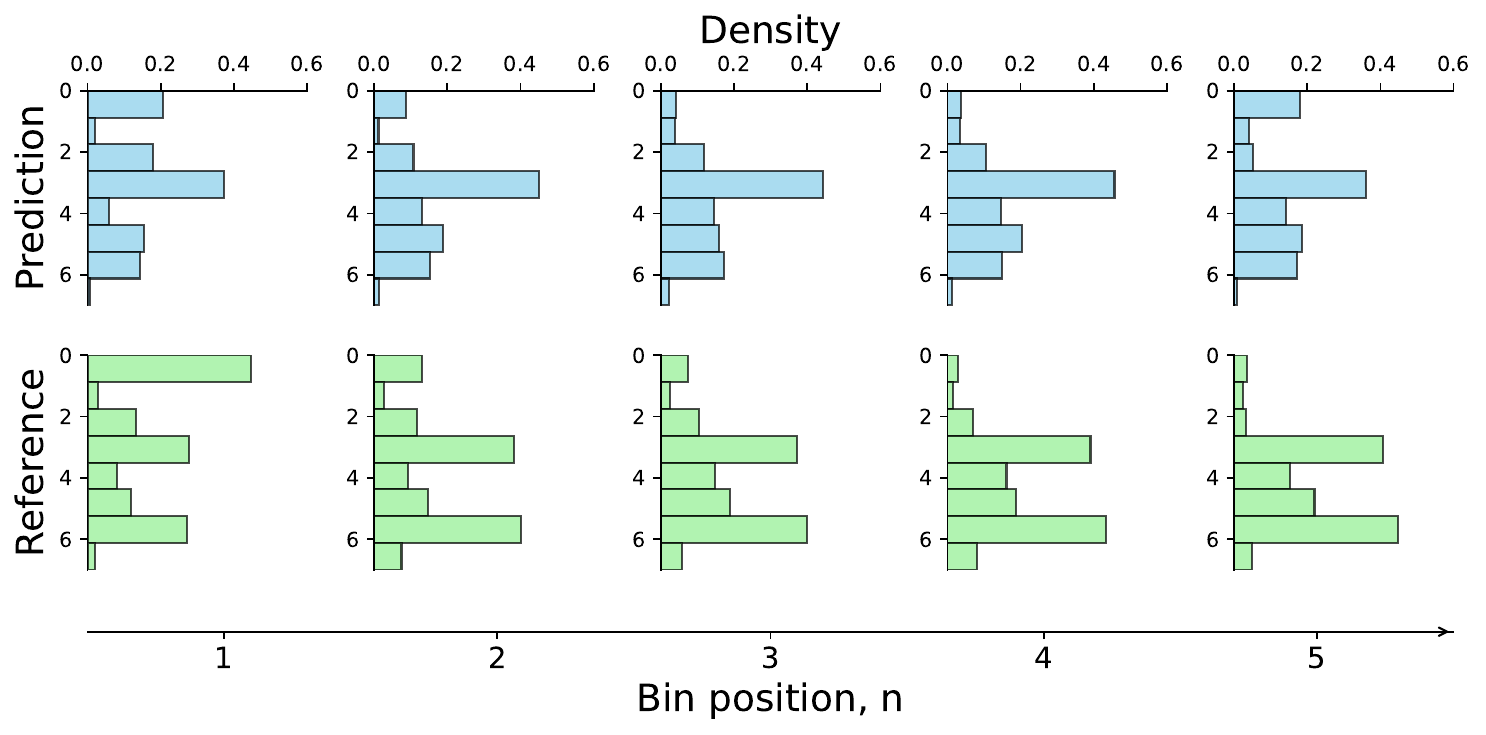}  
    \caption{Positional discourse distribution comparisons~(N=5). Top row: The discourse distribution of model predictions on News Discourse test set (Llama2-7b, finetuned on Kaggle All the News). Bottom row: Test set reference distributions.}
    \label{fig:discourse distribution}
    \vspace{-2mm}
\end{figure}

Despite the fluency achieved by (large) Language Models, they struggle to organize discourse structures like humans.
In Fig.~\ref{fig:discourse distribution}, we observe disparities between the discourse distributions of model predictions and human-written references when the News articles are divided into 5 positional bins.
To quantitatively capture these gaps, we introduce the Positional Discourse Divergence (PDD), denoted as $D_{pos}$, as an automatic metric. 
Equation~\ref{equ:pos_div} outlines the calculation for applying PDD to compare a \emph{predicted} article against its corresponding \emph{reference}:
\begin{equation}\label{equ:pos_div}
    D_{pos} = \frac{1}{N} \sum_{n=1}^{N} D_{KL}(p^n(r)+\bm{\epsilon}||q^n(r)+\bm{\epsilon})      
\end{equation}

Firstly, both articles are segmented into $N$ positional bins.
Note the number of bin, $N$, should be smaller than the number of sentences in both the reference and the prediction. We denote $p^n(r)$ to represent the distribution of discourse role $r$ for the \emph{reference} in the $n$-th position bin, and $q^n(r)$ to represent the distribution for the \emph{generated} article.
These discourse distributions are calculated by the frequency density of the discourse roles within each bin.
These discourse distributions are calculated by the frequency density of the discourse roles within each bin.
Then, for each bin $n$, the KL divergence between the discourse distributions is calculated.
\footnote{
Due to the asymmetry nature of KL divergence, $D_{KL}(P||Q)$ is interpreted as the information divergence of Q against P. Accordingly, we employ $q^n(r)$ to denote the discourse distributions of the predictions and $p^n(r)$ for the reference.
}
To address the sparsity in the discourse distribution of a single article, small-value terms, denoted as 
$\bm{\epsilon}$, are introduced. This addition helps in avoiding instances of zero probabilities in the distribution.

To compute PDD or other discourse measurements like exact match, it is inevitable to employ a discourse role classifier for labeling both prediction and reference articles.
An off-the-shelf discourse classifier, trained on human-annotated data with a defined schema (e.g., the News Discourse dataset for news domain), can serve this purpose. 
Further information regarding the discourse classifiers is provided in Appendix~\ref{sec:result_cls}.

\subsection{Interpreting the Metric}
\subsubsection{Set vs. Individual Predictions}
Much like the BLEU score, the Positional Discourse Divergence can be applied to a set of text, including both the set of model predictions and the set of reference articles. 
The underlying assumption is that all articles within a given set adhere to similar discourse structures, for example, being News articles of the same sports genre. 
Consequently, the discourse distributions of this set of articles offer a more accurate estimate of the discourse distribution specific to that genre.

In the assessment of a single predicted article against a reference set, the focus is on how well the article aligns with the target genre. In contrast, when comparing a prediction set against a reference set, the evaluation exams the model's overall ability to generate content of that specific genre.

\subsubsection{Bin Number}

\begin{figure}
    \centering
    \includegraphics[scale=0.5]{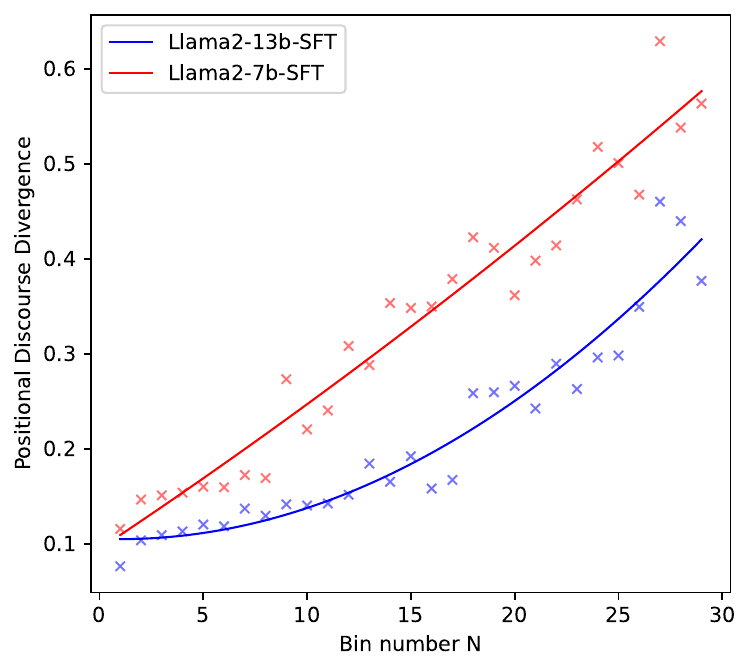}  
    \caption{Positional Discourse Divergence vs. Bin number ($N$) for predictions by two language models on the News Discourse test set. Training details in Appendix~\ref{appen:sft}. Curves represent best-fit quadratic curves.
    }
    \label{fig:divergence_vs_bin}
    \vspace{-2mm}
\end{figure}

The bin number, $N$, plays a crucial role in determining the sensitivity of PDD to local variations, as illustrated in Fig.~\ref{fig:divergence_vs_bin}.
Therefore, the behavior of PDD varies with the choice of $N$. 
Intuitively, a larger $N$ implies lower tolerance for local perturbations.
When $N$ equals the number of sentences, the PDD is essentially equivalent to the exact match metric.
Whereas, when $N$ equals 1, it describes the overall discourse role distribution gaps between the prediction to the reference.

To illustrate, we fine-tuned Llama2-7b and Llama2-13b~\citep{touvron2023llama} and compared their predictions with the reference News articles.
The details of the supervised fine-tuning process are explained in Appendix~\ref{appen:sft}. 
The PDD curves, illustrating the performance with different choices of bin numbers, are presented in Fig.~\ref{fig:divergence_vs_bin}. The gaps in performance are effectively captured by the disparities between the two PDD curves.

\begin{table*}[ht]
\centering  %
\renewcommand{\arraystretch}{1.2}
   \scalebox{0.9}{
        \begin{tabular}{lccclccc}
        \hlinewd{0.8pt}
        \multicolumn{1}{c}{\multirow{2}{*}{Metrics}} & \multicolumn{3}{c}{Human}         &  & \multicolumn{3}{c}{GPT-4}                                                                     \\ \cline{2-4} \cline{6-8} 
        \multicolumn{1}{c}{}                         & News Discourse & LFQA & Recipe1M+ &  & \multicolumn{1}{c}{News Discourse} & \multicolumn{1}{c}{LFQA} & \multicolumn{1}{c}{Recipe1M+} \\ \hline
        Exact Match  &  0.26  &  0.29  &  0.43  &  &  0.42  &  0.24  &  0.25  \\
        ROUGE-L  &  0.30  &  0.24  &  0.39  &  &  0.44  &  0.19  &  0.26  \\
        BLEU  &  0.24  &  0.31  &  0.46  &  &  0.48  &  0.28  &  0.32  \\
        BertScore  &  \textbf{0.45}  &  0.42  &  0.63  &  &  0.63  &  \textbf{0.42}  &  0.68  \\ \hline
        PDD  &  0.42  &  \textbf{0.49}  &  \textbf{0.66}  &  &  \textbf{0.65}  &  0.38  &   \textbf{0.71}  \\ \hlinewd{0.8pt}
        \end{tabular}
    }
    \caption{Cohen's Kappa with human and GPT-4 coherence evaluations. For News Discourse, bin number $N=8$ was used, while for LFQA and Recipe1M+, $N=3$. Human evaluation involves 50 randomly selected example pairs for each dataset, while GPT-4 evaluation uses 300 pairs.}
    \label{tab:human}
\vspace{-2mm}
\end{table*}

\section{Metric Validation}

To validate the efficacy of the Positional Discourse Divergence metric, we evaluate its agreement with human assessments, and GPT-4 on article coherence. 
Additionally, we compare PDD against baseline automatic metrics, such as exact match, BLEU, and BertScore. To assess generalizability, we conduct this validation across three different domains, each characterised by distinct human annotated, sentence-level discourse schemas:

\textbf{(I) News.}
We utilize the News Discourse dataset~\citep{choubey-etal-2020-discourse}, comprising 802 documents across four genres and three media sources. 
The average number of sentences per article is 14.6. 
Manual annotations for the News Discourse dataset follow the theory of functional discourse schema proposed by \citet{van1988news, van2013news}. 
This schema defines discourse based on eight types of relations between each sentence and the main event.

\textbf{(II) Long-form QA.}
Long-Form Question Answering (LFQA) involves providing comprehensive answers composed of multiple sentences. 
\citet{xu-etal-2022-answer} proposed an discourse ontology of six sentence-level functional roles also following the theory of functional  discourse structure.
The discourse annotations are collected on three recent LFQA datasets (ELI5~\citep{fan-etal-2019-eli5}, WebGPT~\citep{nakano2022webgpt}, and Natural Questions~\citep{10.1162/tacl_a_00276}). 
A total of 640 answer paragraphs were released, with an average of 6.1 sentences per paragraph.

\textbf{(III) Recipes.}
We adopt the discourse schema proposed by \citet{liu-etal-2022-plug} which includes seven discourse roles based on cooking actions specifically designed for recipes.
They annotated the Recipe1M+ dataset~\citep{moryossef-etal-2019-step, marin2021recipe1m+} with a rule-based annotation system following the proposed schema.
The Recipe1M+ contains over 1M textual recipes and ingredients.

For further information regarding dataset details and schema definitions, please refer to Appendix~\ref{appen:schema}.

\subsection{Comparison with Other Metrics}
We validate the effectiveness of our metric, PDD, by assessing its inter-agreement with human evaluations and GPT-4 coherence evaluations. 
The human evaluation setup details can be found in Appendix~\ref{appen:human}.
In our comparison, PDD is evaluated alongside several automatic metrics, including exact match, BLEU~\citep{papineni2002bleu}, ROUGE-L~\citep{lin2002manual} and BertScore~\citep{zhang2019bertscore}. 
Notably, only PDD and exact match focus on directly measuring discourse structure, while the others are designed for assessing n-gram or semantic similarity.

As rating long-form articles with absolute scores is a relatively complicated and subjective task, we instead ask evaluators compare two perturbed variations of the original reference article.
Cohen's Kappa is computed between the metrics and evaluators based on these preference annotations.
We create two variations in the way that prevents resulting PDD values from exhibiting a heavy left-tail issue and ensuring a more accurate kappa estimation. 
In Variation 1, we randomly shuffle all the sentences, whereas in Variation 2, we initially segment the article into a randomly selected number of bins and then shuffle sentences only within their respective positional bins.

In Tab.~\ref{tab:human}, we report the Kappa with both human and GPT-4 coherence evaluations. The details of the prompt and survey templates are shown in Appendix~\ref{appen:templates}.  
Our PDD metric demonstrates consistent good agreement (0.4-0.6) in News Discourse and LFQA, achieving substantial agreement (>0.7) on Recipe1M+ dataset. 
The notable performance on the Recipe1M+ dataset can be attributed to the strong order-dependent nature of recipes:  
A shuffled question-answer format may be challenging to understand, but a disordered recipe is nearly incomprehensible.

Another observation on News Discourse dataset, indicates Kappas with human evaluations are generally lower than those with GPT-4.
This discrepancy is likely due to the much longer length of news articles compared to question answering and recipe datasets, posing a more challenging task for human readers.

Our PDD metric significantly outperforms baseline metrics of Exact Match, Rouge-L and BLEU, while achieving comparable Kappa with the BertScore. 
We attribute the good performance of BertScore to its ability in carrying textual knowledge from the pre-trained BERT. %
In our experiment setup, both Variation 1 and 2 are shuffled from the same articles.
Consequently, the metrics based on n-gram and semantic similarities can effectively distinguish examples closer to the original version and therefore achieve high kappa values.
However, when comparing the discourse structure between two different articles of the same genre, they are likely to have very different n-gram or semantic content while maintaining a similar discourse structure.
In this case, only Exact Match and PDD can capture the divergence between discourse structure.

\subsection{Discussion}
Our experimental findings yield the following noteworthy observations:
\begin{itemize}[noitemsep,topsep=1pt]
    \itemsep 0em
    \item The behavior of PDD, as indicated by the formula in Eq.~\ref{equ:pos_div}, converges towards Exact Match as the chosen bin number increases.
Conversely, with a smaller value of $N$, PDD consistently outperforms Exact Match in terms of kappa. 
This observation validates our initial hypothesis that permitting a certain level of local variation does not detrimentally impact the overall reader experience.
    \item PDD consistently exhibits high kappa scores across diverse domains, emphasizing the significance of preserving discourse structure in text across various subject areas.
    \item PDD is specifically designed to evaluate the underlying discourse structure. 
It is not only simple and model-free, eschewing reliance on pre-trained language models, but also interpretable because of its intrinsic use of KL divergence. 
\end{itemize}

\section{Conclusion}
In conclusion, the exploration of text generation with natural underlying structure remains a significantly under-explored domain. 
Addressing this gap, we introduced PDD, a simple and model-free metric designed to assess discourse structure. 
By quantifying the divergence between discourse distributions within position bins, PDD exhibits robust agreement with human and GPT-4 coherence evaluations across three representative domains, outperforming a range of baseline metrics. 
Our hope is that PDD will stimulate future research endeavors focused on unraveling the intricacies of underlying structure in text generation.

\section*{Limitations}
\paragraph{Discourse classifier requirement} 
We note that our PDD requires a discourse classifier when applied to model predictions.
Although this necessity is inevitable in evaluating the discourse structure alignment for any other metric such as Exact Match, it underscores the dependence on annotated data with the target discourse schema for training.

\paragraph{Choice of bin number $N$}
The choice of bin number will affect the performance of the PDD. 
However, the ideal choice of $N$ may vary for different articles: The optimal number of sections the article should be segmented into.
While trends may exist within specific genres or datasets, in general, it requires certain level of domain expertise to determine the optimal bin number.

\bibliography{anthology,custom}
\bibliographystyle{acl_natbib}

\appendix

\section{Background and Related Works}
\subsection{Discourse Structure} \label{appen:discourse structure}

Discourse structure investigates the organization of language into larger units like paragraphs, sections, and complete articles. 
In this work, we focus on the communicative functions within entire articles served by those linguistic units.
Therefore, texts from different domains are characterized by different discourse schemas, as their linguistic units also play different functional roles.
The discourse roles of scientific papers or experimental abstracts~\citep{liddy1991discourse,mizuta2006zone} include background, methodology, experiments and findings.
In the domain of long-form question answering \citet{xu2022we}, the discourse function of each sentence can be answer, summary, example and so on. 
\citet{liu-etal-2022-plug} developed a discourse schema for recipes based on actions and controlled the generation process according to the predicted discourse sequences. 
The explicit functional discourse structure of news reports was addressed and leveraged~\citep{van2013news, choubey-etal-2020-discourse,liu2023instructsctg} by defining roles based on their relations with the main event, such as consequence and journalist evaluation.

Multiple established frameworks also proposed different definition of discourse structure, which focus on how each linguistic unit relates to each other through discourse connectives, such as causal, temporal, etc.
For instance, Rhetorical Structure Theory, RST \citep{mann1988rhetorical}, seeks to identify rhetorical relations between text segments and form a hierarchical organization of discourse.
The Penn Discourse Treebank, PDTB \citep{prasad2008penn}, defines its schema based on low-level discourse connectives presented in the text.

\subsection{NLG Metrics}
Traditional NLG metrics, such as BLEU~\citep{papineni2002bleu} and ROUGE-L~\citep{lin2002manual}, measure lexical n-gram overlaps to assess fluency, but they have limitations in capturing semantic similarity.
Later works tried to improve the hard lexical matching with soft word embedding matching~\citep{ng-abrecht-2015-better} or stemming and synonym matching~\citep{lavie-agarwal-2007-meteor}.

Recently, by leveraging contextual embeddings from BERT~\citep{kenton2019bert}, a series of metrics can successfully capture semantic similarity with references or even textual quality without references\citep{zhao-etal-2019-moverscore,zhang2019bertscore,yuan2021bartscore}.
However, as BERT is argued that can only capture limited discourse information~\citep{koto-etal-2021-discourse,laban-etal-2021-transformer,beyer-etal-2021-incoherence}, they are not suitable for evaluating the discourse structure in long texts.

DiscoScore~\citep{zhao-etal-2023-discoscore} is a BERT-based metric, specifically designed to model local discourse coherence for summarization and document-level machine translation tasks. 
By leveraging Center theory~\citep{grosz-etal-1995-centering}, they modelled discourse similarity by focus frequency and transitions.
Recently LLMs have recently been utilized as judges to evaluate various aspects of text quality, such as coherence and fluency. For example, the PairS framework \citep{liu2024aligning} employs LLMs to assess and compare the quality of generated text.

\section{Discourse Classifier}\label{sec:result_cls}
A discourse classifier is usually a lightweight language model trained on sentence-discourse role pairs. Here we report the classifier performance achieved:

For the News domain, we train a DistilBERT~\citep{sanh2019distilbert} as the discourse role classifier on the News Discourse training set and evaluated on the validation set using human-annotated gold labels. The classifier achieves an accuracy of $67\%$.

In the Recipe domain, the reported performance of the discourse role classifier, by \citet{liu-etal-2022-plug}, achieves an accuracy of $92\%$.
It was a a DistilBERT model trained on the Recipe1M+ training set and evaluated on the validation set using silver annotations generated by a rule-based system.

For LFQA, \citet{xu-etal-2022-answer} achieved an accuracy of $54\%$ by a T5-large model, which shows comparable performances to human.
The classifier was trained and tested on the ELI5 dataset.

\section{LLM SFT Details}\label{appen:sft}
We fine-tuned two language models, the 8-bit LoRa versions of Llama2-7b and Llama2-13b, using Kaggle All the News train set comprising 42.4K samples after filtering. 
The models receive news headlines as input and aim to generate the corresponding news articles. 
Training employed consistent hyper-parameters: a learning rate of $3 \times 10^{-4}$, 2 epochs, LoRa parameters $r=8$, $\alpha=16$, and dropout set at $0.05$. 
The models were trained on a single RTX a6000 48GB, requiring 12 and 23 hours for Llama2-7b and Llama2-13b, respectively.

\section{Human Evaluation Details} \label{appen:human}
The human evaluation was conducted using Amazon Mechanical Turk (MTurk). We obtained three preference annotations for each example pair from native English-speaking crowd workers. The final results were determined based on the majority preference among the three evaluations. Crowd workers were compensated at a rate of 15 pounds per hour for their participation in the evaluation process.

\section{Discourse Schema}
\label{appen:schema}
The definition of the discourse schema we used for news articles:
\begin{itemize}[noitemsep,topsep=1pt]
    \itemsep 0em
    \item \textbf{Main Event}: The major subject of the news article.
    \item \textbf{Consequence}: An event or phenomenon that is caused by the main event.
    \item \textbf{Previous Event}: A specific event that occurred shortly before the main event.
    \item \textbf{Current Context}: The general context or world state immediately preceding the main event.
    \item \textbf{Historical Event}: An event occurring much earlier than the main event.
    \item \textbf{Future Consequences}: An analytical insight into future consequences or projections.
    \item \textbf{Journalist Evaluation}: A summary, opinion or comment made by the journalist.
    \item \textbf{Anecdotal Event}: An event that is uncertain and cannot be verified. The primary purpose is to provide more emotional resonance to the main event.
\end{itemize}

The definition of the discourse schema for LFQA:
\begin{itemize}[noitemsep,topsep=1pt]
    \itemsep 0em
    \item \textbf{Organizational sentence}: An organizational sentence is to inform the reader how the answer will be structured.
    \item \textbf{Answer summary}: An answer sentence that plays a summary role, which can often suffice by themselves as the answer to the question.
    \item \textbf{Answer}: Answer sentences which explain or elaborate on the summary.
    \item \textbf{Example}: The example provided in answers, which discussed a particular entity or concept that is different from the rest of the answer sentences.
    \item \textbf{Auxiliary Information}: Provide information that are related, but not directly asked in the question.
    \item \textbf{Miscellaneous}: Various other roles that shows up in human answers, such as the limitation of the answer or the source of the answer.
\end{itemize}

\noindent The definition of the discourse schema we used for recipes:
\begin{itemize}[noitemsep,topsep=1pt]
    \itemsep 0em
    \item \textbf{Pre-processing} means the preparations of ingredients or cooker.
    \item \textbf{Mixing} includes actions of combining one or more ingredients together.
    \item \textbf{Transferring} is for the actions of moving or transferring food or intermediate food to a specific place.
    \item \textbf{Cooking} represents the actual cooking actions, which could vary drastically across different recipes.%
    \item \textbf{Post-processing} usually refers to the following up actions after the `cooking' stage, such as `cooling down', `garnish'.
    \item \textbf{Final} refers to the last few actions before serving the food or the serving action itself.
    \item \textbf{General} includes the rest of actions which cannot be classified into the above categories.
\end{itemize}

\section{Data Preprocessing}
\label{appen:preprocess}
For News Discourse, we filtered the dataset based on the following conditions:
\begin{itemize}[noitemsep,topsep=1pt]
    \itemsep 0em
    \item Containing special characters: @, [, +.
    \item Having total number of words over 800 or below 100.
    \item Containing random comments.
    \item Containing more than two reports.
\end{itemize}
Then we pre-process the data by
\begin{itemize}[noitemsep,topsep=1pt]
    \itemsep 0em
    \item Removing extra space.
    \item Removing reporting source.
    \item Removing journalist names.
    \item Removing emoji.
\end{itemize}

\noindent For Recipe1M+, we filter it based on the following codintions:
\begin{itemize}[noitemsep,topsep=1pt]
    \itemsep 0em
    \item Containing irrelevant information, such as advertisements, reviews and comments.
    \item Having total number of words over 300 or below 50.
    \item Duplicate recipes.
\end{itemize}

\noindent For LFQA, we filtered out all model generated answer paragraphs, because they contain sentences that do not have assigned discourse roles.

\section{Evaluation Templates} \label{appen:templates}

\paragraph{Human evaluation instruction}
Below, we provide the instruction example for human evaluation on the News Discourse dataset, where evaluators were directed to express their preference. The instructions for LFQA and Recipe1M+ are similar, with certain domain-specific keywords substituted, such as News headline becoming Recipe title.

\enquote{
    Read the two versions of the news for the given headline and rank their coherence following the guideline below.
    
    Coherence guidelines:
    
    1. Flow of Sentences: Evaluate how well the sentences transition from one to another. A fluent text should have seamless connections between sentences.
    
    2. Logical Organization: Evaluate how well the sentences are organized and the ideas are conveyed. A coherent text should have a clear and precise structure.
    
    General guidelines:
    
    1. Be Objective: Please focus on the coherence of writing, not the content or opinions expressed.
    
    2. Please rate which one is preferred between the two versions. 
    
    News headline: \{headline\}
    
    Version 1:
    \{version1\}
    
    Version 2:
    \{version2\}
}

\paragraph{GPT-4 evaluation prompt}
Below, we provide the prompt template for GPT-4 coherence evaluation on the News Discourse dataset.
Although the GPT-4 was instructed to rate with scores, but the scores are converted to preference later.
The instructions for LFQA and Recipe1M+ are similar, with certain domain-specific keywords substituted, such as News headline becoming Recipe title.

\enquote{
        Pretend you are a human reader. Please evaluate the coherence of the two given news articles.  
        Guideline:  
        
        1. Rate on a scale of 1 to 10, where 1 represents very low coherence, and 10 indicates very high coherence.
        
        2. Consider the flow of ideas and the ordering of sentences. A highly coherent article should have a better sentence ordering. 
        
        3. Must return ratings in JSON format only: \{"score1": [your rating for version 1], "score2": [your rating for version 2]\}

        News headline:
        [ headline ]

        News version 1:
        [ version1 ]

        News version 2:
        [ version2 ]

        Rating:\
    }

\end{document}